\documentclass[conference]{IEEEtran}

    \IEEEoverridecommandlockouts
    \usepackage{cite}
    \usepackage{amsmath,amssymb,amsfonts}
    \usepackage{algorithmic}
    \usepackage{graphicx}
    \usepackage{textcomp}
    \usepackage{xcolor}
    \usepackage{listings}
    \usepackage{color}
    \usepackage{import}

    \lstloadaspects{formats}

\definecolor{dkgreen}{rgb}{0,0.6,0}
\definecolor{gray}{rgb}{0.5,0.5,0.5}
\definecolor{mauve}{rgb}{0.58,0,0.82}

\colorlet{punct}{red!60!black}
\definecolor{background}{HTML}{EEEEEE}
\definecolor{delim}{RGB}{20,105,176}
\colorlet{numb}{magenta!60!black}

\lstdefinelanguage{json}{
    basicstyle=\fontsize{7}{8}\selectfont\ttfamily,
    numbers=left,
    numberstyle=\scriptsize,
    stepnumber=1,
    numbersep=3pt,
    showstringspaces=false,
    breaklines=true,
    frame=lines,
    backgroundcolor=\color{background},
    literate=
     *{0}{{{\color{numb}0}}}{1}
      {1}{{{\color{numb}1}}}{1}
      {2}{{{\color{numb}2}}}{1}
      {3}{{{\color{numb}3}}}{1}
      {4}{{{\color{numb}4}}}{1}
      {5}{{{\color{numb}5}}}{1}
      {6}{{{\color{numb}6}}}{1}
      {7}{{{\color{numb}7}}}{1}
      {8}{{{\color{numb}8}}}{1}
      {9}{{{\color{numb}9}}}{1}
      {:}{{{\color{punct}{:}}}}{1}
      {,}{{{\color{punct}{,}}}}{1}
      {\{}{{{\color{delim}{\{}}}}{1}
      {\}}{{{\color{delim}{\}}}}}{1}
      {[}{{{\color{delim}{[}}}}{1}
      {]}{{{\color{delim}{]}}}}{1},
}
    \def\BibTeX{{\rm B\kern-.05em{\sc i\kern-.025em b}\kern-.08em
        T\kern-.1667em\lower.7ex\hbox{E}\kern-.125emX}}
    
\begin{document}

    \title{Second Hand Price Prediction for Tesla Vehicles}

    \author{\IEEEauthorblockN{Sayed Erfan Arefin}
    \IEEEauthorblockA{
    Dhaka, Bangladesh \\
    erfanjordison@gmail.com}

    }

    \maketitle

    \begin{abstract}
    The Tesla vehicles became very popular in the car industry as it was affordable in the consumer market and it left no carbo foot print. Due to large decline in the stock prices of Tesla Inc. at the beginning of 2019, Tesla owners started selling their vehicles in the used car market. This used car prices depended on attributes such as, model of the vehicle, year of production, miles driven and the battery used for the vehicle. Prices were different for a specific vehicle in different months. In this paper, it is discussed how a machine learning technique is being implemented in order to develop a second hand Tesla vehicle price prediction system. To reach this goal, different machine learning techniques such as decision trees, support vector machine (SVM), random forest and deep learning were investigated and finally was implemented with boosted decision tree regression. I future, it is intended to use more sophisticated algorithm for a better accuracy.

\end{abstract}

\renewcommand\IEEEkeywordsname{Keywords}

    \section{Introduction}
An American car manufacturing company Tesla Inc was founded on 2003. It is one of the companies that developed electric cars to a manufacturing standard and made it available for the consumer market \cite{ref:about-tesla-wikipedia}. In the beginning of 2019, the stock price of Tesla started to drop. This made Tesla vehicle owners feared and most of them started selling of their vehicle. A huge demand increased in the used car market for Tesla vehicles. But the price for Tesla vehicle depended on the stock price of the vehicle and it impacted largely in the used Tesla vehicle market. \\
Tesla vehicles highly depends on Tesla company’s post sells services. In order to keep the vehicles usable in the market and also decline of the stock prices the accurate prediction of used car prices were very important for the new and existing customers and the company itself. \\
Predicting the price of a used Tesla vehicle became very important and also challenging. The whole process of making a second hand price prediction system consisted two sections. \\
One section predicts the price of current tesla vehicles owned by a customer. It helps suggesting different driving habits which will minimize the price drop of the vehicle.
Another section was to determine the price of Tesla vehicle by monitoring the prices of used cars from different websites that provided sells information of used Tesla vehicles. We will be focusing on how this second section worked. Become it is the most important part of the whole system. \\
The websites that provide this information includes truecar.com, autotrader.com, cars.com etc. We used a data scrapper to retrieve this information and feed it to our machine learning system in order to predict the price. The prices of a same vehicle were different in a range of three months (January, February and March of 2019). This means, the month of recorded price of a certain Tesla vehicle was also very important. The attributes that was found in the datasets were: Model, Year, Battery, Price, Miles, Exterior color, Interior type, Wheel type, Spoiler type etc. Since the prices were getting very saturated, four major attributes were focused to predict the price of Tesla vehicles in order to have an accurate result. These are: Model, Year, Battery, Price, Miles.
In this paper how the datasets were used to implement to develop a prediction system using Microsoft Azure Machine Learning studio will be discussed. We will also discuss the test out of three different machine learning techniques that were primarily chosen to predict the price: Random forest, Boosted decision tree and K-nearest neighbor. Later the Boosted decision tree was selected as the main algorithm of the price prediction system, due to less errors.

    \section{Background study}
Tesla shares faced their steepest price drop in 2019 when the company reported a large un expected loss. Around the same time their co-founder and Chief Technology Officer (CTO) JB Strubel departed from his executive ranks. The stock price dropped 13.6\% and was \$228.82 at the close. This was the largest drop since the share prices fell 13.9\% on September 28, 2018. Later the stock price dropped more than 14\% which is the steepest drop of price since 2013. \cite{ref:cnbc-tesla-1} \\
Even though prices of stock were declining, the company increased their affords to prevent this. Some analysis suggested a similar decline of stock price was observed of a popular Tech company Netflix Inc. Later which recovered from such a situation. Eddie Yoon, founder of think and tank EddieWouldGrow, suggested Tesla’s stock price in 2019 is very similar to Netflix’s stock prices in 2011. \cite{ref:cnbc-tesla-2-similar-to-netflix}\\
It was very important for the company and also the customers to predict the price of second hand Tesla vehicles. The prices of a same vehicle were different in a range of three months (January, February and March of 2019). This means, as the time passed by prices of Tesla vehicle of a same type changed. Which is a very important factor in our machine learning technique to be used.

    \section{Dataset} \label{sec:dataset}
This section includes dataset collection methods and daaset description. In sub-section \ref{dataset-collection} how the datasets were collected are discussed and in sub-section \ref{dataset-descrition} data set description can be found. The attributes of the datasets and analysis of the dataset can be found in this sub-section.

\subsection{Dataset collection}
\label{dataset-collection}
Tesla vehicle users sell their vehicles and this information can be retrieved from some well recognized websites. This includes truecar.com, autotrader.com, cars.com etc. There is another tool available online which is import.io. Using this tool data can be scrapped or extracted from the mentioned websites. 3 months of data of used cards were collected from cars.com, truecar.com and autotrader.com. Using import.io data was scrapped and datasets were collected in csv (Comma separated values) formatted files. Approximately total of 1600 entries were collected. 

\subsection{Description of Dataset}
\label{dataset-descrition}
The dataset is labeled properly. There are 6 attributes of the dataset which are order independent and they are shown in table xyz. A sample of the dataset can be observed in table \ref{dataset-sample}.  Attributes are described as follows.

    \begin{description}
        \item[$\bullet$] \textbf{ Model:} This is the model of Tesla vehicle. There are four models available. They are, Model S, Model X, Roadstar 2dr and Model 3.
        \item[$\bullet$] \textbf{	Year:} This indicates the manufacturing year.
        \item[$\bullet$] \textbf{	Battery:} Battery indicates the battery model used for that specific vehicle. Battery is very important for a Tesla vehicle. Battery degradation, makes the battery to hold less charge which puts the vehicle in a bad condition. That also effects the price. 
        \item[$\bullet$] \textbf{	Price:} The price on which the vehicle was sold. The price is on US dollar.
        \item[$\bullet$] \textbf{	Miles:} The total miles the vehicle has driven. 
    \end{description}
    
    \begin{figure*}
        \includegraphics[width=\linewidth]{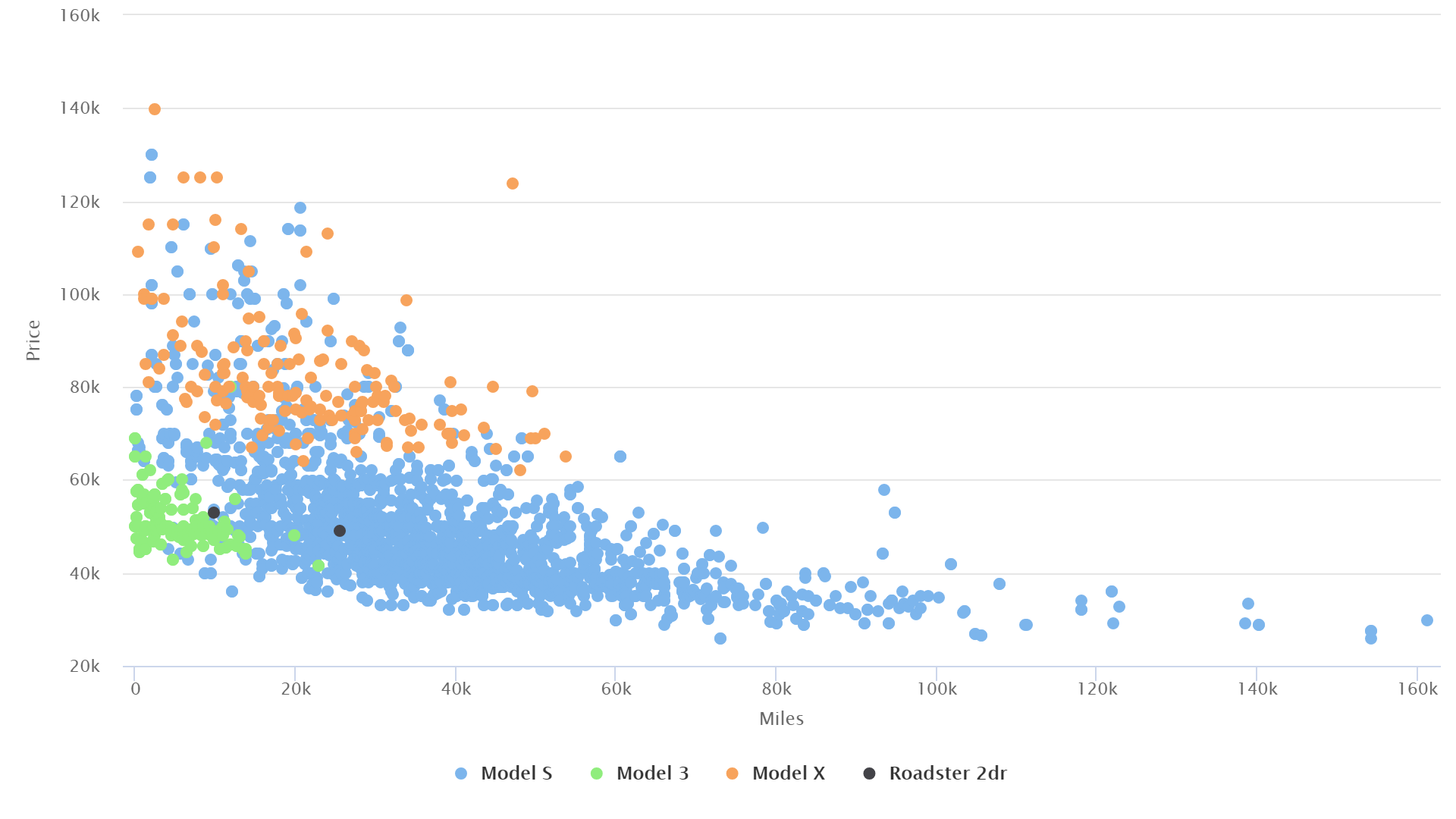}
        \caption{Boosted Decision Tree Implementation in Machine Learning Studio}
        \label{fig:data-visual-2}
    \end{figure*}

    \begin{figure*}
        \includegraphics[width=\linewidth]{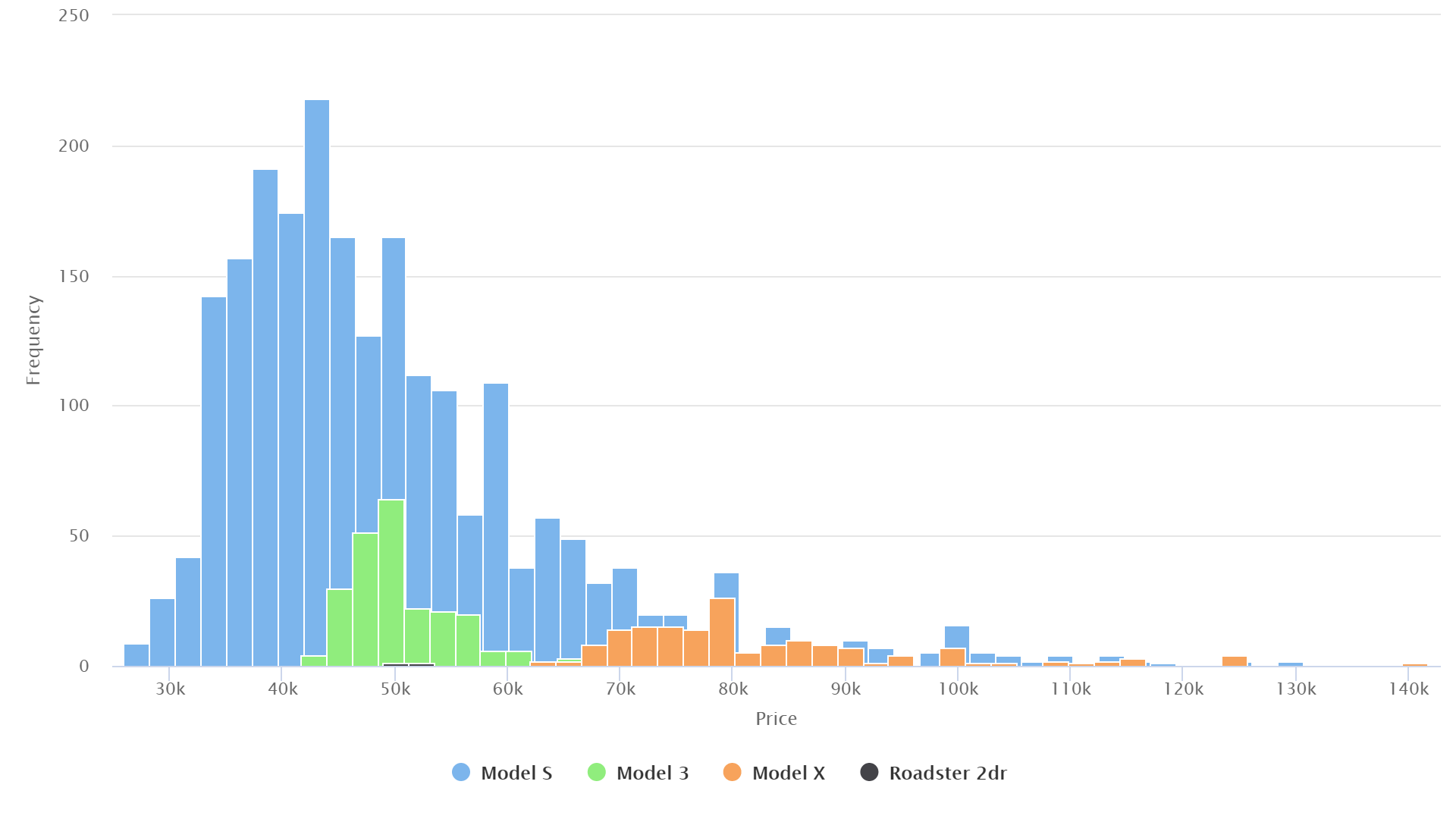}
        \caption{Scatter plot of the dataset: Price vs Frequency, colors indicating different models.}
        \label{fig:data-visual-1}
    \end{figure*}

    Rapid miner is a tool available for free, which can be used to pre process datasets, generate insights of the dataset and also run some machine learning techniques, such as deep learning, decision trees, support vector machine etc. \cite{ ref:rapid-miner}. Using this the following scatterplot and histogram were generated.\\
A scatterplot was plotted figure \ref{fig:data-visual-1}. In this graph, X-axis indicates the miles and Y-axis indicates price. The Models are marked with different colors. \\
It can be seen that, most of the Model 3 vehicles were sold before it reached a 20,000 miles drive by their owners and are sold at a price range of 40,000 USD to 70,000 USD. Average of Model 3 sold was 51,332.67 USD. For Model S, it can be observed that it has been sold in a vast range of driven miles, having large distribution in terms of mile, the average price it was sold is, 49,430.64 USD. Most of the Model X, were sold before it reached a 60,000 miles drive by the owners. The average price it was sold is, 83,475.01 USD which is much higher than any other models of Tesla vehicles. In case of Roadstar 2dr, the samples were very low. Average price they were sold is, 51,493 USD. \\
	
A histogram was generated figure \ref{fig:data-visual-2}. In this graph, X-axis indicates the prices and Y-axis indicates frequency. The Models are marked with different colors. 
It can be observed that, the highest number of model sold was Model S, having a wide distribution in price. Compare to Model S, Model 3 and Model X are sold at a lower frequency. Price distribution of Model 3 is between 40,000 USD and 70,000 USD. Model X are sold over 60,000 USD. Roadstar 2dr had a minimum sale and is priced around 51,493 USD.\\setminus
The mean, median and mode of the whole dataset can be found in table \ref{table:mean-median-mode} and model based average prices can be found in table \ref{table:model-average-price}.

    \begin{table}
        \centering
        \caption{THE DATA ATTRIBUTES }
        \label{dataset-attributes}
        \begin{tabular}{|l|l|l|} 
            \hline
            Serial & AttributeTitle & DataType  \\ 
            \hline
            1      & Model          & String    \\ 
            \hline
            2      & Year           & String    \\ 
            \hline
            3      & Battery        & String    \\ 
            \hline
            4      & Price          & Number    \\ 
            \hline
            5      & Miles          & Number    \\ 
            \hline
        \end{tabular}
    \end{table}
    
    \begin{table}
        \centering
        \caption{Dataset Sample}
        \label{dataset-sample}
        \begin{tabular}{|l|l|l|l|l|} 
            \hline
            Model   & Year & Battery & Price & Miles  \\ 
            \hline
            Model S & 2013 & Base    & 34200 & 36800  \\ 
            \hline
            Model 3 & 2018 & 75      & 46995 & 2193   \\ 
            \hline
            Model S & 2018 & 75D     & 64900 & 1095   \\ 
            \hline
            Model X & 2016 & P90D    & 84984 & 20680  \\ 
            \hline
            Model S & 2016 & 75D     & 58989 & 20303  \\
            \hline
        \end{tabular}
    \end{table}

    \begin{table}
        \centering
        \label{table:mean-median-mode}
        \caption{Dataset Mean, Median and Mode}
        \begin{tabular}{|l|l|l|} 
        \hline
        Mean & Median         & Mode            \\ 
        \hline
        51041.84~ ~ ~ ~ ~                                & 46800~ ~ ~ ~ ~ & 39999~ ~ ~ ~ ~  \\
        \hline
        \end{tabular}
    \end{table}
    \begin{table}
        \centering
        \caption{Average price based on models}
        \label{table:model-average-price}
        \begin{tabular}{|l|l|l|l|l|} 
        \hline
        Models        & Model 3  & Model S  & Model X  & Roadstar 2dr  \\ 
        \hline
        Average Price & 51332.67 & 49430.64 & 83475.01 & 51493         \\
        \hline
        \end{tabular}
    \end{table}

    \section{Implementation}

As the dataset is labeled properly, it is considered to be used for supervised learning. In order to find out the best machine learning technique, different machine learning techniques were tested and based on the RMSE further decision was taken.\\
The following machine learning techniques were considered in the experiment.

\subsection{Support Vector Machine (SVM)}
Gegic, E., Isakovic, B. showed that in case of using datasets for Ford and Volkswagen the Support Vector Machine (SVM) showed an accuracy of 48.23\% which was higher than Artificial Neural Network algorithm \cite{ref:svm-ann-journal}. In case of predictiong Tesla Prices SVM was also applied.

\subsection{Deep Learning}
Peerun, Saamiyah \& Henna Chummun showed that a neural network which had one hidden layer with two nodes produced the smallest mean absolute error, among different machine learning techniques that were experimented. \cite{ref:deep-learning-ann}. Though they also showed that SVM performed slightly well. They used data sets from Ford, Mazda, Opel, Renault, Toyota, Citreon prices in their paper. \\
Thus, in case of predicting the price of Tesla vehicles it will also be used in the experiment.

\subsection{Random Forest}
N. I. Nwulu showed that in case of oil price predictions, random forest performed well compare to decision tress, it had lower RMSE \cite{{ref:oil-price-decision-tree-random-forest} }. \\
Pudaruth, Sameerchand also showed that in case of car price prediction with the dataset of Nissan and  Toyota Random forest algorithm performed well compare to Decision Tree algorithm \cite{ref:knn-randomforest-deision-tree}. \\
In Tesla price prediction Random forest algorithm was also used.

\subsection{Decision Tree}
Pudaruth, Sameerchand build decision trees with Nissan and Toyota cars prices as datasets \cite{ ref:knn-randomforest-deision-tree}. They used decision tree to predict the prices and had a success rate of 84\%. Thus, decision tree is also used in this experiment to predict Tesla Vehicle prices.

\subsection{Boosted Decision Tree}
\begin{figure}
  \includegraphics[width=\linewidth]{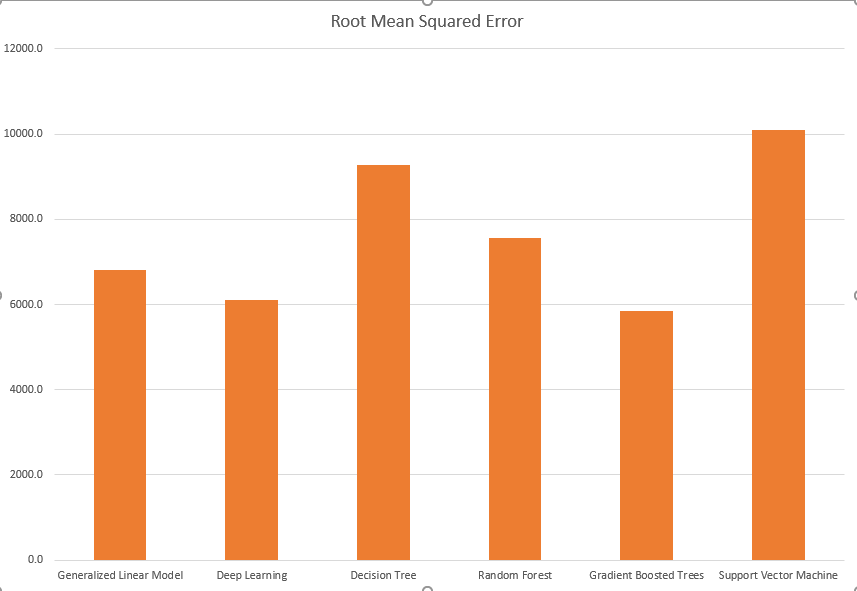}
  \caption{Results of Rapid Miner: RMSE}
  \label{fig:rapid-miner-results}
\end{figure}

In order to produce a prediction model, Grading boosting is a machine learning technique for regression and classification problems. This technique is mainly used for decision trees. The model is built in a stage-wise approach, similar to other boosting methods and it generalized by optimization of an arbitrary differentiable loss function. \\
Gradient boosted trees had performed well in the Rapid Miner experiment. It can be seen in the results from Rapid Miner containing the RMSE figure \ref{fig:rapid-miner-results}. \\
Boosted Decision Tree Regression means, the decision tree is formed based on boosting. This means, each tree is dependent on prior trees. Preceded tree residual is fitted in the current tree and this is how the algorithm learns. \\
Thus Boosted Decision Tree Regression was also considered for the experiment.

\subsection{Error comparison}
RMSE also known as Root Mean Square Error, is a good way to evaluate recommender systems. RMSE penalizes an algorithm more incase of wrong results. Thus, it was considered for deciding which Machine learning technique to be used for the prediction system. In table \ref{rapid-miner} the RMSE results of different machine learning techniques are given, that was generated using the Tesla vehicle datasets.\\
It can be seen that Gradient Boosted Trees had a RMSE of 5849.8. But for Boosted Decision Tree Regression implementation on Machine Learning Studio, shows a RMSE of 4969.53 table \ref{errors-from-evaluate-model}. Which is slightly better than Gradient Boosted Trees. \\
Thus Boosted Decision Tree Regression was considered for the final implementation of the prediction system.

\subsection{Final Implementation}
\begin{figure}
  \includegraphics[width=\linewidth]{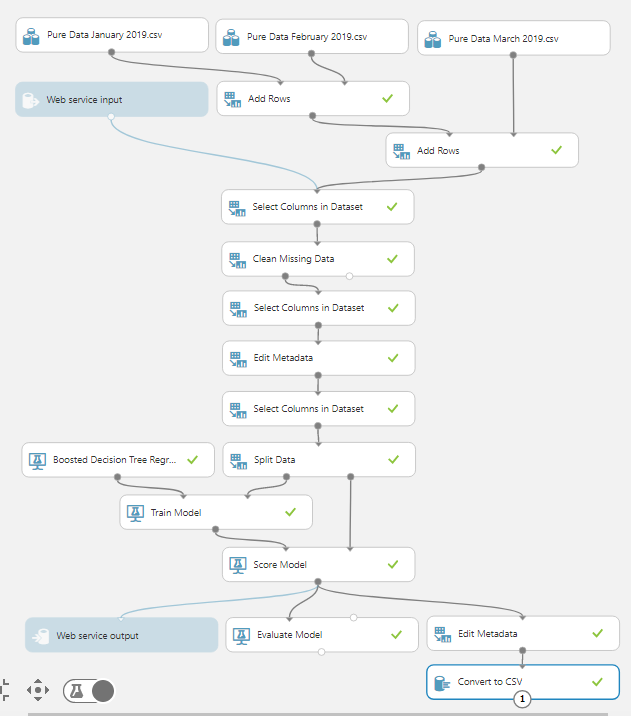}
  \caption{Boosted Decision Tree Implementation in Machine Learning Studio}
  \label{fig:machine-learning-model}
\end{figure}

\begin{figure}
  \includegraphics[width=\linewidth]{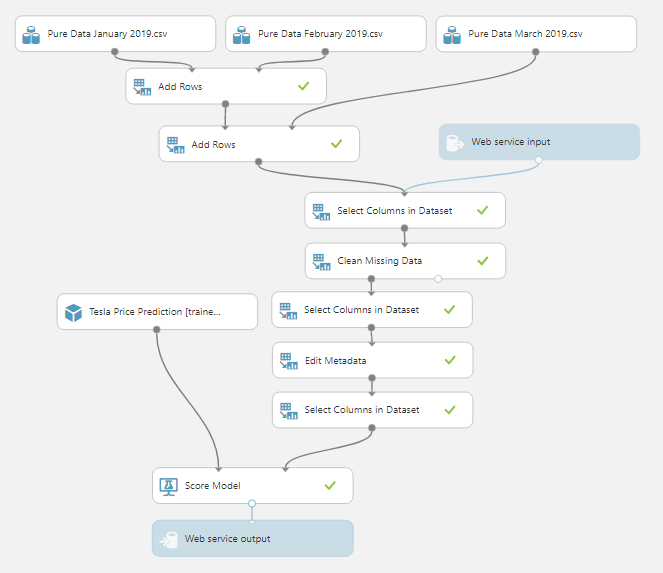}
  \caption{Machine learning web-service with the trained model}
  \label{fig:machine-learning-trained-model}
\end{figure}

As the boosted decision tree regression shows the lowest RMSE, it was decided to be used in the machine learning API. Machine Learning Studio is provided by Microsoft Azure \cite{ref:machine-learning-studio}. In this platform models can be trained by uploading datasets, different machine learning techniques can be defined Later the trained data can be used to deploy a webservice to be consumed by other backend services. This has a modular interface. \\
In the implementation first, the datasets of different months were uploaded and combined into one dataset, then the columns discussed in section \ref{sec:dataset} (Model, Year, Battery, Price, Miles) were selected. There were some rows with missing data. Those were cleaned up before being used in the training. After that, Boosted Decision Tree Regression was executed and it provided a scored model. Web service input and web service output was also used but those had no purpose in this step, but need to be added for creating the web service later on. From the scored module the data can be converted to CSV and later was downloaded. A snapshot of the data is given in Table \ref{scored-model-snapshot}. The whole implementation can be visualized in figure \ref{fig:machine-learning-model}. \\
Modules used for implementing the Boosted Decision Tree Regression are given below with explanations.
\begin{description}
  \item[$\bullet$] \textbf{	Add Rows: }This module take two datasets as inputs and merges them. The columns of the datasets need to be same. The output contains the merged dataset.
 \item[$\bullet$] \textbf{	Select Columns in Dataset:} This module helps select columns of the dataset provided in the input and outputs a dataset with the selected columns.
 \item[$\bullet$] \textbf{	Clean Missing Data} This module cleans all the rows which have missing values.
 \item[$\bullet$] \textbf{	Edit Metadata:} With this, types of the attributes can be changed or corrected in case of the attributes not being presented properly. 
 \item[$\bullet$] \textbf{	Split Data: } With this module the set can be split into two datasets based on the provided ratio. In this case the ratio was 75\% and 25 \%. The split was randomized. The first output gives the larger split and the second output provide the second half of the split.
 \item[$\bullet$] \textbf{	Boosted Decision Tree Regression:} This module along with the Train Model module runs the Boosted Decision Tree Regression training. The algorithm is pre implemented with the module. 
 \item[$\bullet$] \textbf{	Train Model:} This takes in the module which defines the algorithm to use and the second input takes in the dataset that will be used for training. This is only used for regression and classification problems. So, it is suitable for this implementation.
 \item[$\bullet$] \textbf{	Score Model: }This generates a score or predicted value. In this case, a sample of the scored model is given in table \ref{scored-model-snapshot}.
 \item[$\bullet$] \textbf{	Evaluate Model:} This generates Mean Absolute Error  (MAE), Root Mean Squared Error  (RMSE), Relative Absolute Error, Relative Squared Error, Coefficient of Determination. In this case, the results from evaluate model are given in table \ref{errors-from-evaluate-model}
 \item[$\bullet$] \textbf{	Web service input:} This module takes input when deployed as web service. 
 \item[$\bullet$] \textbf{	Web service output:} This module provide response to web service input when deployed as web service.
 \item[$\bullet$] \textbf{	Convert to CSV:} This converts the input into a CSV file, which can be downloaded for further reference.
\end{description} 
It took 26 seconds to run the experiment. Because this is hosted in a cloud environment in this case, Microsoft Azure, it took very little time to run. After completing the experiment, it was turned into a webservice, where the Web service input and Web service output were used along with the trained model. This can be seen in figure \ref{fig:machine-learning-trained-model}. \\
This exposed a web service API, which later consumed by the back end. This API is discussed in section \ref{sec:testing}, while testing. 

\begin{table}
  \centering
  \caption{Errors from Evaluate Model}
  \label{errors-from-evaluate-model}
  \begin{tabular}{|l|l|} 
  \hline
  Error & Value                                            \\ 
  \hline
  Root Mean Squared Error                          &  4969.534921  \\ 
  \hline
   Relative Absolute Error                          &  0.297097     \\ 
  \hline
   Relative Squared Error                            &  0.098898     \\ 
  \hline
   Coefficient of Determination                      &  0.901102     \\ 
  \hline
   Mean Absolute Error                               &  3478.598936  \\
  \hline
  \end{tabular}
\end{table}

\begin{table}
  \centering
  \caption{Scored Model snapshot}
  \label{scored-model-snapshot}
  \begin{tabular}{|l|l|l|l|l|l|}  \hline
  Model   & Year & Battery     & Price & Miles & Scored Labels  \\ \hline
  Model S & 2013 & Base        & 34200 & 36800 & 36950.03516    \\ \hline
  Model X & 2017 & 90D         & 88470 & 12229 & 88367.42969    \\ \hline
  Model 3 & 2018 & 75          & 46995 & 2193  & 52845.39063    \\ \hline
  Model X & 2017 & P100D       & 98500 & 33959 & 110876.7656    \\ \hline
  Model S & 2014 & Performance & 44855 & 44980 & 44208.11328    \\ \hline
  Model X & 2018 & 75D         & 84995 & 1379  & 83715.78906    \\ \hline
  Model X & 2016 & P90D        & 81500 & 25250 & 79317.74219    \\ \hline
  Model S & 2016 & 90          & 59855 & 23010 & 52286.07031    \\ \hline
  Model S & 2017 & P100        & 98000 & 18871 & 104973.0547    \\ \hline
  Model S & 2013 & 60          & 43995 & 15559 & 39556.19922    \\ \hline
  Model S & 2015 & 70D         & 47995 & 47367 & 42450.09766    \\ \hline
  Model S & 2016 & 90D         & 53995 & 55370 & 54899.95703    \\ \hline
  Model 3 & 2018 & 75          & 54900 & 1254  & 52441.74609    \\ \hline
  Model S & 2012 & Performance & 36995 & 38030 & 42229.39844    \\ \hline
  \end{tabular}
\end{table}

\begin{table}
  \centering
  \caption{Rapidminer results for different machine learning techniques}
  \label{rapid-miner}
  \begin{tabular}{|l|l|l|} 
  \hline
  Model                    & \begin{tabular}[c]{@{}l@{}}Root Mean \\Squared Error\end{tabular} & \begin{tabular}[c]{@{}l@{}}Standard \\Deviation\end{tabular}  \\ 
  \hline
  Generalized Linear Model & 6812.9                                                            & 383.2                                                         \\ 
  \hline
  Deep Learning            & 6110.2                                                            & 346.0                                                         \\ 
  \hline
  Decision Tree            & 9267.3                                                            & 740.2                                                         \\ 
  \hline
  Random Forest            & 7557.1                                                            & 542.7                                                         \\ 
  \hline
  Gradient Boosted Trees   & 5849.8                                                            & 341.4                                                         \\ 
  \hline
  Support Vector Machine   & 10108.7                                                           & 708.1                                                         \\
  \hline
  \end{tabular}
\end{table}

    \section{Testing} \label{sec:testing}
\begin{figure}
    \includegraphics[width=\linewidth]{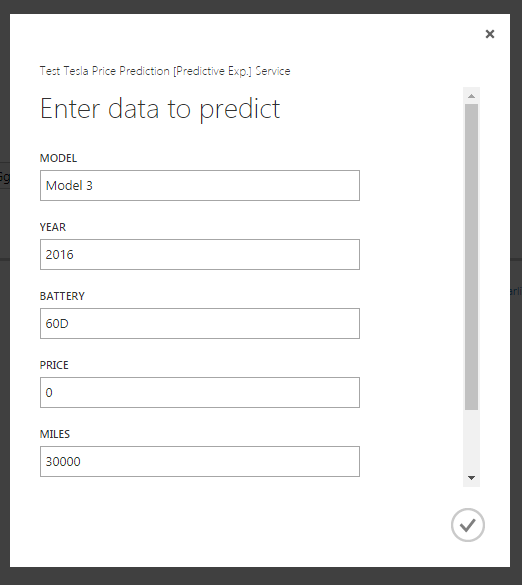}
    \caption{Web service input on Machine learning studio}
    \label{fig:machine-learning-studio-web-service}
\end{figure}

Testing the implemented API is a very important step. It will be tested if the price is predicted properly.  Also, it will be tested if the machine learning webservice is working properly. \\
In order to test if the price is predicted properly, the API is fed with Used vehicle prices randomly. A small piece of code that is used to run a loop. Each time of the loop, a request has been made to the API with the entries from the April dataset. The predicted price is listed with the corresponding entry. The predicted price is subtracted from the original price in order to find the error margin. It showed a RMSE to the RMSE found in the evaluate model errors shown in table \ref{errors-from-evaluate-model}
To run this piece of code, firstly it must me tested if the machine learning API is working properly. Web services in the Machine Learning Studio includes an input form to test out the webservice. It was tested and worked properly. The request form is shown in figure \ref{fig:machine-learning-studio-web-service}. \\

\begin{figure}
    \includegraphics[width=\linewidth]{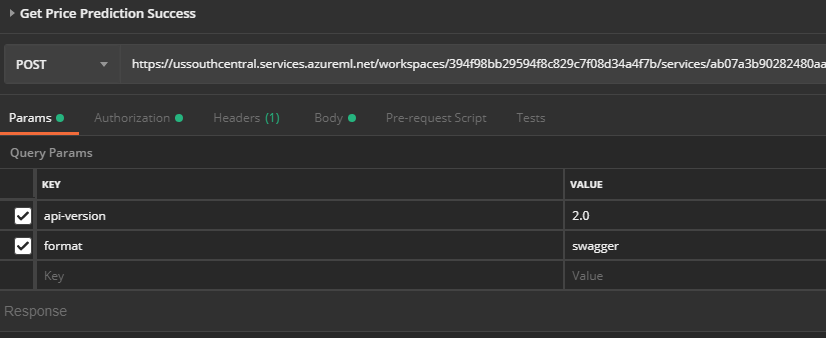}
    \caption{Testing with Postman: parameters for request.}
    \label{fig:testing-with-postman-request-parameters}
\end{figure}

\begin{figure}
    \includegraphics[width=\linewidth]{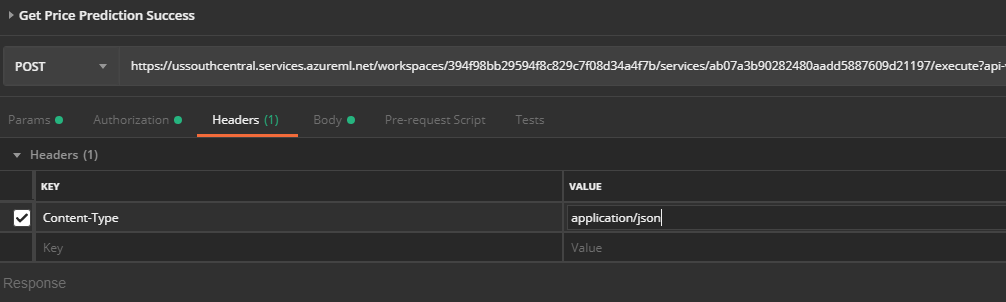}
    \caption{Testing with Postman: Headers for request.}
    \label{fig:testing-with-postman-request-headers}
\end{figure}

To test the API from another source, Postman will be used. Postman is a tool that helps create a Rest API request using all the necessary requirements such as providing the request with header, parameters, authorization tokens etc. \cite{ref:about-postman}. To test out the web services API with Postman parameters, headers and authorization is populated with the values in table \ref{table:postman-parameters}, \ref{table:postman-header}, \ref{table:postman-authorization} subsequently, which were retrieved from the web services API. It is also shown in figure \ref{fig:testing-with-postman-request-parameters}, \ref{fig:testing-with-postman-request-headers} subsequently. The request is finally populated with the following JSON (Java Script Object Notation) which included all the parameters required.

\begin{lstlisting}[language=json,firstnumber=1]
    {
        "Inputs": {
            "input1":
            [
                {
                    "Model": "Model X",   
                    "Year": "2017",   
                    "Price": "0",
                    "Battery": "75",   
                    "Miles": "19000",
                    "Date": "2019-01-01"
                }
            ]
        },
        "GlobalParameters":  {
        }
    }
\end{lstlisting}

The response from the API is shown below. It is received also as JSON.

\begin{lstlisting}[language=json,firstnumber=1]
    {
        "Results": {
            "output1": [
                {
                    "Model": "Model X",
                    "Year": "2017",
                    "Battery": "75",
                    "Price": "0",
                    "Miles": "19000",
                    "DateCreated": "1/1/2019 12:00:00 AM",
                    "Scored Labels": "81789.6640625"
                }
            ]
        }
    }
\end{lstlisting}

This shows that the web service API is functioning properly. Then the error calculation step was executed.

\begin{figure}
    \includegraphics[width=\linewidth]{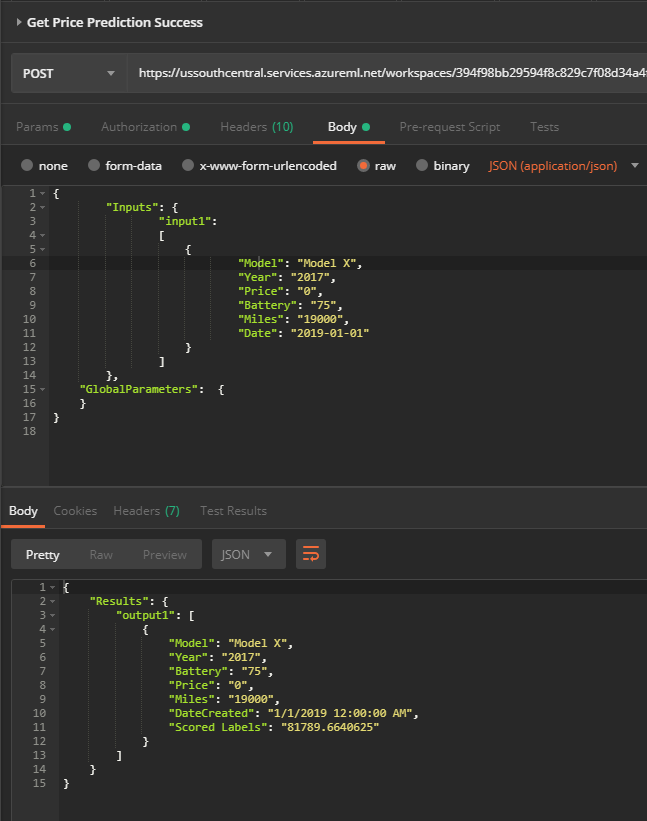}
    \caption{Testing with Postman: Request Body and Response}
    \label{fig:testing-with-postman-request-response}
\end{figure}

\begin{table}
    \centering
    \caption{Postman parameters}
    \label{table:postman-parameters}
    \begin{tabular}{|l|l|}
    \hline
    Key          & Value                                             \\ \hline
    api-version  & 2                                                 \\ \hline
    format       & swagger                                           \\ \hline
    \end{tabular}
\end{table}

\begin{table}
    \centering
    \caption{Postman headers }
    \label{table:postman-header}
    \begin{tabular}{|l|l|}
    \hline
    Key          & Value                                             \\ \hline
    Content-Type & application/json                                  \\ \hline
    \end{tabular}
\end{table}

\begin{table}
    \centering
    \caption{Postman authorization }
    \label{table:postman-authorization}
    \begin{tabular}{|l|l|}
    \hline
    Key          & Value                                             \\ \hline
    Bearer Token & \{Retrieved from web service portal, kept secret\} \\ \hline
    \end{tabular}
\end{table}
    \section{The Final System}
The final system was developed using the MERN stack. MERN stands for Mongo DB, Express Js, React Js and Node Js. The backend of the system was developed on Typescript. User needs to sign up in order to use the system and provide the VIN of the vehicle. \\
VIN is vehicle identification number. Which is unique for all the vehicles, similar to mac addresses that is being used in case of network devices. From VIN a vehicle model, year of production, manufacturer name etc can be extracted. Thus, it was used as an entry point to the system. With this, and the prediction web service, the system showed a predicted price for that specific vehicle.\\
Improvements on the final system is considered for future work.

    \section{Conclusion}
Predicting the price of a used vehicle can be challenging. In this paper, a price prediction system for Tesla vehicles was discussed. Different machine learning techniques were performed with the collected dataset in order find the lowest RMSE. Techniques considered were: Deep Learning, Support Vector Machine, Boosted Decision Tree Regression, Random Forest and Decision Trees. Boosted Decision Tree algorithm showed the lowest RMSE. It was later implemented in Machine Learning Studio and with the trained model a web service was deployed. Which was later used in the main back end of the prediction system. In the future it is intended to improve the accuracy of the prediction.

\end{document}